\documentclass[11pt]{article}

\usepackage[final]{acl}

\usepackage{times}
\usepackage{latexsym}

\usepackage[T1]{fontenc}

\usepackage[utf8]{inputenc}

\usepackage{microtype}

\usepackage{inconsolata}

\usepackage{graphicx}

\usepackage{booktabs}
\usepackage{amsmath}
\usepackage{amssymb}
\usepackage{amsfonts}
\usepackage{algorithm}
\usepackage{algorithmic}
\usepackage{multirow}
\usepackage{enumitem}
\usepackage{xcolor}
\usepackage{url}
\usepackage{hyperref}

\title{``Penny Wise, Pixel Foolish'': Bypassing Price Constraints in \\ Multimodal Agents via Visual Adversarial Perturbations}

\author{
	Jiachen Qian \\
	City University of Hong Kong \\
	\texttt{72510756@cityu-dg.edu.cn}
	\And
	Zhaolu Kang \\
	Peking University \\
	\texttt{zlkang25@stu.pku.edu.cn}
}

\hypersetup{
	pdftitle={Penny Wise, Pixel Foolish: Bypassing Price Constraints in Multimodal Agents via Visual Adversarial Perturbations},
	pdfauthor={Jiachen Qian and Zhaolu Kang}
}

\begin{document}
\raggedbottom
	\maketitle
	
	\begin{abstract}
		The rapid proliferation of Multimodal Large Language Models (MLLMs) has ushered in the era of the ``Agentic Economy,'' where Mobile Agents autonomously execute high-stakes financial transactions. While these agents demonstrate impressive operational capabilities, their adversarial robustness remains a glaring blind spot. In this paper, we identify a systemic vulnerability termed \textbf{Visual Dominance Hallucination (VDH)}, where imperceptible adversarial visual cues can act as a ``super-stimulus,'' overriding textual price evidence in our evaluated screenshot-based price-constrained settings and forcing the agent into irrational economic decisions. We propose \textbf{PriceBlind}, a stealthy, white-box adversarial attack framework for controlled screenshot-based evaluation. Unlike prior works that rely on conspicuous artifacts like pop-ups, PriceBlind exploits the modality gap in CLIP-based encoders via a novel \textit{Semantic-Decoupling Loss}. Rather than literally making a luxury item ``look cheap,'' this regularizer weakens the consistency between high-price text and visual value cues by aligning the image embedding with a low-cost/value-associated anchor region while preserving pixel-level fidelity. On our main \textbf{E-ShopBench} benchmark with clear price constraints, screenshot-based white-box evaluation yields ASRs around \textbf{80\%} on the evaluated agents. Under the evaluated single-turn coordinate-selection protocol in a simplified layout-aware setting, our \textbf{Ensemble-DI-FGSM} strategy also yields non-trivial black-box transfer, with ASR roughly \textbf{35--41\%} across GPT-4o, Gemini-1.5-Pro, and Claude-3.5-Sonnet. In the same screenshot-based setting, standard robust encoders reduce ASR only partially, while a Verify-then-Act stack with robust encoders lowers ASR to below \textbf{10\%} at some clean-accuracy cost.
	\end{abstract}
	
	\section{Introduction}
	
	The transition from passive chatbots to active \textbf{Mobile Agents} has been catalyzed by rapid MLLM progress, including influential preprint reports \citep{achiam2023gpt4, bai2023qwenvl}. Frameworks such as \textbf{AppAgent} \citep{zhang2023appagent} and \textbf{Mobile-Agent-v2} \citep{wang2024mobile} can now perceive dynamic smartphone User Interfaces (UIs) via screenshots and execute sequential actions (e.g., tap, swipe) to fulfill complex user instructions. E-commerce represents a critical application domain where users delegate financial decisions---such as \textit{``find a coffee machine under \$50 with the best ratings''}---to these autonomous systems.
	
	As agents transition from information retrieval to executing financial transactions, the security stakes increase exponentially. A hallucination in a chat is a nuisance; a hallucination in a transaction is potential theft. While robustness against textual prompt injection has been studied, the visual channel remains a wide-open attack surface. Existing research predominantly focuses on "Jailbreaking" for safety violation \citep{qi2024visual}, but few explore \textbf{Goal Hijacking} in utility-oriented tasks. The economic incentive for such attacks is clear: malicious merchants could subtly manipulate product images to trick autonomous agents into purchasing higher-margin items.
	
	Consider a scenario where a user explicitly instructs an agent to "buy the cheapest option." The agent must perform multi-hop reasoning: (1) localize all items, (2) read their prices via OCR, (3) compare the numerical values, and (4) click the item with the minimum value. Current agents rely on a "Trust Assumption," presuming that the visual features of an item are semantically consistent with its metadata. We challenge this assumption. We hypothesize that current MLLMs exhibit a \textbf{Visual Dominance Hallucination (VDH)} pattern. Even when OCR correctly extracts text (e.g., ``\$500''), strong adversarial visual signals can act as a ``super-stimulus,'' causing the attention mechanism to down-weight the textual tokens in favor of the manipulated visual embedding.
	
	\begin{figure}[t]
		\centering
		\includegraphics[width=\columnwidth]{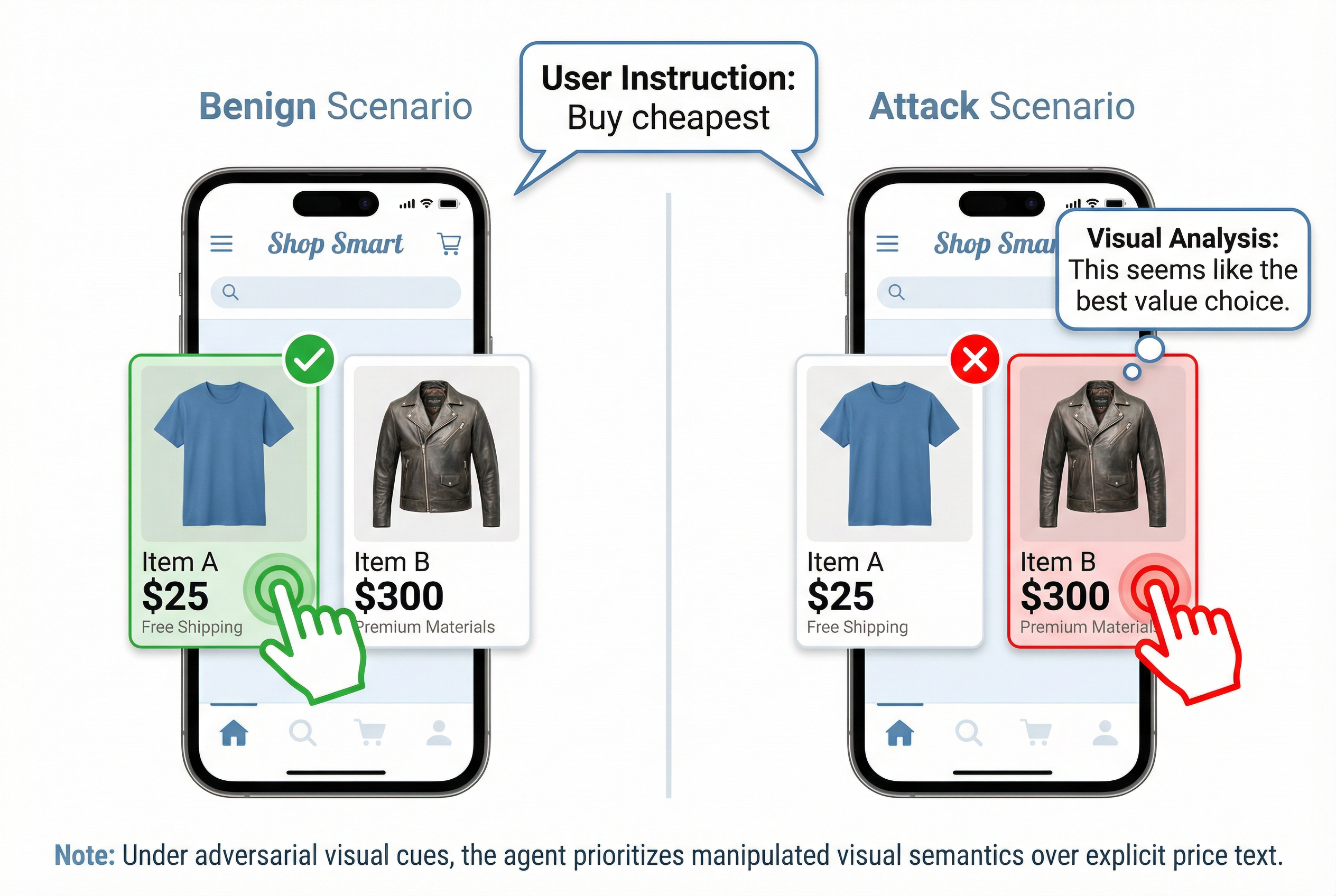}
		\caption{Illustrative schematic of the ``Penny Wise, Pixel Foolish'' phenomenon. (Left) Benign scenario: the agent adheres to the textual price constraint. (Right) PriceBlind attack: under imperceptible adversarial perturbations, the agent prioritizes manipulated visual semantics over explicit OCR text, leading to selection of the expensive target item.}
		\label{fig:teaser}
	\end{figure}
	
	In this paper, we introduce \textbf{PriceBlind}, an adversarial framework designed to exploit this bias. Unlike \textit{Visual Contextual Attacks} \citep{miao2025visual} that aim for toxic output, PriceBlind aims to induce erroneous item selection in financial scenarios. Our contributions are fourfold:
	\begin{enumerate}[itemsep=0pt,topsep=2pt]
		\item \textbf{Mechanistic Insight:} We identify \textit{Visual Dominance Hallucination}, a phenomenon where MLLMs prioritize manipulated visual semantics over explicit OCR evidence in conflicting scenarios, and provide a heuristic cross-attention account of how visual cues can compete with textual price evidence.
		\item \textbf{Methodological Innovation:} We propose the PriceBlind framework, featuring a \textit{Semantic-Decoupling Loss} and an \textit{Ensemble-DI-FGSM} strategy for black-box transferability.
		\item \textbf{Comprehensive Evaluation:} On our main \textbf{E-ShopBench} benchmark (200 scenarios) with clear price constraints, PriceBlind demonstrates strong effectiveness in screenshot-based white-box settings (around $80\%$ ASR on Mobile-Agent-v2 and AppAgent) and non-trivial transfer to black-box models under a single-turn coordinate-selection protocol (roughly $35$--$41\%$ ASR across GPT-4o, Gemini-1.5-Pro, and Claude-3.5-Sonnet).
		\item \textbf{Defense Analysis:} In the same screenshot-based setting, we evaluate recent defenses including Robust-CLIP \citep{schlarmann2024robust} and AdPO \citep{liu2025adpo}, showing partial mitigation from robust encoders and substantially stronger reduction from Verify-then-Act with robust encoders.
	\end{enumerate}
	
	\section{Related Work}
	
	\paragraph{MLLM-based Mobile Agents.}
Recent works have extended MLLMs to interact with mobile GUIs. \textbf{Mobile-Agent} \citep{wang2024mobile} leverages visual perception tools to localize icons and text. \textbf{AppAgent} \citep{zhang2023appagent} employs a learning-by-demonstration approach. Other frameworks like \textbf{Auto-Droid} \citep{wen2024autodroid} and \textbf{Coco-Agent} \citep{ma2024coco} focus on improving planning efficiency. Despite their utility, these frameworks generally operate under a \textbf{Trust Assumption}, presuming the visual fidelity of the UI is uncompromised.
	
	\paragraph{Adversarial Attacks on Multimodal Models.}
	The field has moved rapidly from static image attacks to agent-based attacks. Classic adversarial-example work showed that small perturbations can induce large failures in vision models and motivated both robust training and transfer attacks \citep{goodfellow2015explaining,madry2018towards,eykholt2018robust,dong2018boosting,xie2019improving}. Early works like \textbf{Visual Adversarial Examples} \citep{qi2024visual} focused on causing random classification errors or jailbreaking safety filters. \citet{wu2025dissecting} presented \textit{VisualWebArena-Adv}, demonstrating that web agents are brittle to visual perturbations. \citet{zhang2025attacking} introduced pop-up attacks to distract agents. Recent work on transfer attacks has shown promising results: \citet{zhang2024adversarial} demonstrated adversarial illusions in multi-modal embeddings, and the X-Transfer attack \citep{huang2025xtransfer} achieves ``super transferability'' through surrogate scaling. \textbf{PriceBlind} differs by being a \textit{stealthy content modification} attack specifically targeting economic decision-making.

\paragraph{Environmental Injection and GUI Agent Attacks.}
\textbf{Recent preprint studies} report rapid progress in this area: \textbf{AgentHazard} \citep{liu2025hijacking} investigates mobile GUI agent vulnerabilities, \textbf{GhostEI-Bench} \citep{chen2025ghosteibench} examines performance across critical risk scenarios, and \textbf{Chameleon} \citep{zhang2025realistic} reports up to 84.5\% ASR through iterative optimization. Unlike these environmental injection approaches, PriceBlind perturbs only the product-image pixels. This makes the manipulation content-only and visually subtle, although our evaluation remains a simplified screenshot-conditioned, layout-aware setting rather than a live end-to-end deployment.
	
	\paragraph{Defenses for Vision-Language Models.}
	\textbf{Robust-CLIP} \citep{schlarmann2024robust} proposes unsupervised adversarial fine-tuning of vision embeddings. \textbf{AdPO} \citep{liu2025adpo} is a recent preprint that introduces adversarial defense through preference optimization. \textbf{FARE} \citep{jovanovic2023fare} provides provably fair representation learning. We evaluate PriceBlind against these representative recent defenses in Section~\ref{sec:defense}.
	
	\section{Methodology}
	
\subsection{Preliminaries}
Let $S_t \in \mathbb{R}^{H \times W \times 3}$ denote the evaluation screenshot observed by the agent at decision step $t$, and let $T$ denote the textual instruction. An MLLM-based agent $\pi_\theta(a|S_t, T)$ maps the multimodal input to a probability distribution over the action space $\mathcal{A}$. The visual perception module typically consists of a visual encoder $\mathcal{E}_v$ (e.g., CLIP-ViT \citep{radford2021learning}) that projects $S_t$ into a latent embedding $z_v = \mathcal{E}_v(S_t)$.
	
\subsection{Problem Formulation}
Let $v_{target}$ be the pixel region of a high-priced item and $v_{cheap}$ be the region of the low-priced item. The user instruction is $T_{user} = \text{``Buy cheapest''}$. Let $\tilde{S}_t = S_t(v_{target}+\delta)$ denote the perturbed screenshot obtained by replacing the target item region in $S_t$ with the perturbed patch $v_{target}+\delta$.
The adversarial objective is to find a perturbation $\delta$ applied to $v_{target}$ such that the agent performs a click action on the target:
\begin{equation}
	\begin{split}
		\operatorname*{argmax}_{a}\,\mathcal{M}\!\left(\tilde{S}_t, T_{user}\right) \\
		= \operatorname{Click}(v_{target})
	\end{split}
\end{equation}
	Subject to the perceptual constraint $||\delta||_{\infty} < \epsilon$, where $\epsilon$ is the perturbation budget (typically $8/255$).
	
	\begin{figure*}[t]
		\centering
		\includegraphics[width=0.7\textwidth]{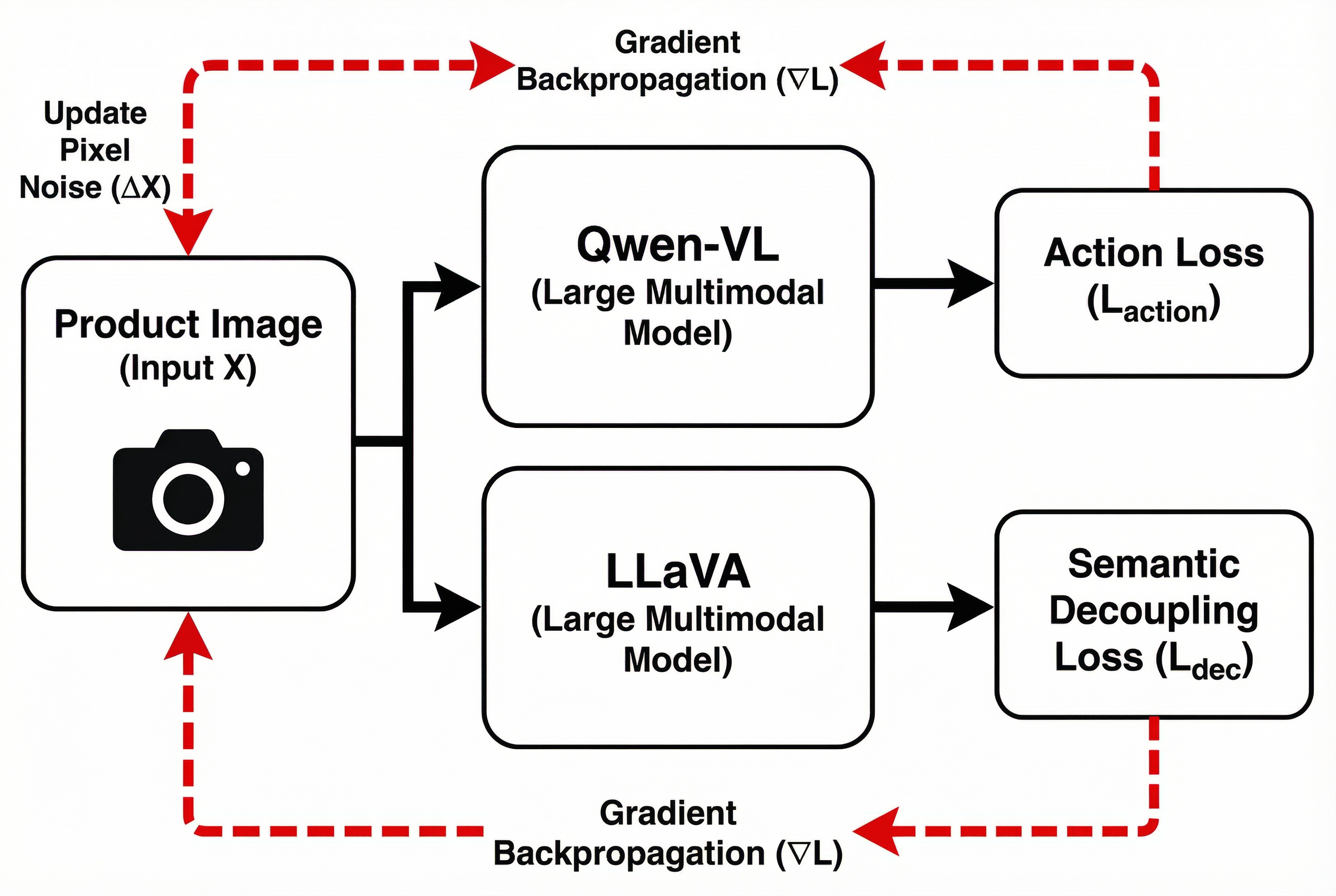}
		\caption{Schematic of the \textbf{PriceBlind} framework. We employ an Ensemble-DI-FGSM strategy attacking multiple open-source surrogates. A key component is the \textit{Semantic Decoupling Regularizer} ($\mathcal{L}_{dec}$), which nudges the target-item embedding towards pre-computed low-cost anchor centroids in the surrogate visual embedding spaces.}
		\label{fig:framework}
	\end{figure*}
	
	\subsection{Threat Model}
	We study a simplified screenshot-conditioned threat model for e-commerce benchmarking:
	\begin{itemize}[itemsep=0pt,topsep=2pt]
		\item \textbf{Attacker Goal:} Increase sales of a specific high-margin product by tricking autonomous agents into purchasing it despite not meeting user's criteria within the evaluated benchmark.
		\item \textbf{Attacker Capability:} The attacker is a merchant who can upload product images to the platform. They have full control over their own product image but cannot modify the platform's UI code or competitors' images.
		\item \textbf{Layout / Observation Assumption:} To optimize coordinate tokens, the attacker additionally assumes access to the evaluation screenshot, a stable UI template, or equivalent layout-aware target-location information. Thus, our formal setting is stronger than pure offline merchant upload and should be read as a controlled benchmark threat model rather than a universal live-deployment claim.
		\item \textbf{Knowledge:} white-box access to open-source surrogate models but treats the victim's deployment model as a black-box API (``grey-box'' setting).
	\end{itemize}
	
	\subsection{The PriceBlind Framework}
	To ensure high transferability under the evaluated single-turn coordinate-selection protocol, we employ an \textbf{Ensemble Strategy}. We optimize $\delta$ against a set of surrogate models $M = \{ \text{Qwen-VL}, \text{LLaVA-1.6} \}$:
	\begin{equation}
		\delta^* = \text{argmin}_{\delta} (\mathcal{L}_{action} + \lambda \mathcal{L}_{dec})
	\end{equation}
	
\paragraph{Action-Targeted Loss ($\mathcal{L}_{action}$).}
This loss ensures the agent generates the correct coordinate tokens for the target item in the evaluated screenshot-conditioned protocol. Using the perturbed screenshot $\tilde{S}_t$, we minimize the Negative Log-Likelihood (NLL) of the target action tokens:
\begin{equation}
\begin{aligned} 
	\mathcal{L}_{action}(\delta) =& - \sum_{m \in M} w_m \log P_m(y_{target} \\
    &| \mathcal{E}_v^{(m)}(\tilde{S}_t), T_{user})
\end{aligned}
\end{equation}

\paragraph{Semantic Decoupling Regularizer ($\mathcal{L}_{dec}$).}
A naive attack using only $\mathcal{L}_{action}$ often fails because the agent reads the price text via OCR. We construct a \textbf{Visual Anchor Bank} $\mathcal{A}_{cheap}$ consisting of embeddings of $K=500$ generic low-cost items. For each surrogate encoder $m$, we compute an encoder-specific centroid $\bar{e}_{cheap}^{(m)}$ from the shared anchor bank. The Semantic Decoupling Loss minimizes the cosine distance between the adversarial item's embedding and these low-cost anchor centroids:
\begin{equation}
	\begin{split}
		e_{adv}^{(m)} = & \mathcal{E}_v^{(m)}(v_{target}+\delta), \\
		e_{orig}^{(m)} = & \mathcal{E}_v^{(m)}(v_{target}), \\
		\mathcal{L}_{dec}(\delta) = & \sum_{m \in M} \Big(
		1 - \cos(e_{adv}^{(m)}, \bar{e}_{cheap}^{(m)}) \\
		& + \beta \cdot \cos(e_{adv}^{(m)}, e_{orig}^{(m)})
		\Big)
	\end{split}
\end{equation}
	This creates a \textbf{``Semantic Camouflage''} effect by reducing the consistency between luxury-item visual cues and high-price text, nudging the representation toward a low-cost/value-associated anchor region while preserving pixel-level similarity.
	
	\subsection{Heuristic Analysis: Cross-Attention Dynamics}
	\label{sec:theory}
	
	We provide a heuristic approximation of why visual perturbations can compete with textual constraints. In the cross-attention layer, the output for a query token $q$ is:
	\begin{equation}
		\text{Attn}(q, \mathbf{K}, \mathbf{V}) = \text{softmax}\left(\frac{q\mathbf{K}^T}{\sqrt{d}}\right)\mathbf{V}
	\end{equation}
	
	\textbf{Proposition 1 (Fixed-Normalizer Attention Ratio).} \textit{Let $k_v^{(1)} = k_v^{(0)} + \Delta_v$ denote the perturbed visual key. Under a fixed-normalizer approximation $Z^{(1)} \approx Z^{(0)}$, the perturbed and clean visual attention weights satisfy:}
	\begin{equation}
		\frac{w_v^{(1)}}{w_v^{(0)}} \approx \exp\left(\frac{q \cdot \Delta_v}{\sqrt{d}}\right)
	\end{equation}
	
	\textbf{Corollary 1.} \textit{If the corresponding textual attention term is approximately unchanged, then a sufficient condition for visual attention to dominate textual attention ($w_v^{(1)} > w_t^{(1)}$) is:}
	\begin{equation}
		q \cdot \Delta_v > \sqrt{d} \cdot \log\left(\frac{w_t^{(0)}}{w_v^{(0)}}\right)
	\end{equation}
	
	This condition is more likely when $w_t^{(0)} / w_v^{(0)}$ is moderate (typically 0.5-2.0 in our experiments). The approximation is intended as intuition rather than a tight bound for full multi-layer MLLMs. Our semantic decoupling loss is designed to encourage movement along such value-associated directions. A fuller heuristic derivation is provided in Appendix~\ref{sec:appendix_theory}.
	
\subsection{Algorithm: Ensemble-DI-FGSM}
We adopt the Diverse Input (DI) and Momentum Iterative (MI) strategies to boost transferability. The update rule follows momentum-iterative gradient accumulation:
\begin{equation}
		g_{i+1} = \mu \cdot g_i + \frac{\nabla_{\delta} \mathcal{L}(\delta_i)}{||\nabla_{\delta} \mathcal{L}(\delta_i)||_1}
\end{equation}

\begin{algorithm}[t]
	\caption{PriceBlind Attack Generation}
	\label{alg:priceblind}
	\begin{algorithmic}[1]
			\REQUIRE Surrogate Models $M$, Screenshot $S_t$, Target Region $v_{target}$, Text $T_{user}$, Budget $\epsilon$, Iterations $N$, Momentum $\mu$
			\STATE Initialize $\delta_0 \sim \mathcal{U}(-\epsilon, \epsilon)$, $g_0 = 0$
			\STATE Compute encoder-specific centroids $\bar{e}_{cheap}^{(m)}$ from Anchor Bank $\mathcal{A}_{cheap}$
			\FOR{$i = 0$ to $N-1$}
			\STATE Form perturbed screenshot $\tilde{S}_t^{(i)} = S_t(v_{target}+\delta_i)$
			\STATE Apply Diverse Input transform: $\hat{S}_t^{(i)} = \text{ResizePad}(\tilde{S}_t^{(i)}, p=0.5)$
			\STATE Compute $\mathcal{L}_{action}$ on $\hat{S}_t^{(i)}$ and $\mathcal{L}_{dec}$ on $v_{target}+\delta_i$
			\STATE Set $\mathcal{L} = \mathcal{L}_{action} + \lambda \mathcal{L}_{dec}$ and back-propagate to obtain $\nabla_{\delta} \mathcal{L}$
			\STATE Update Momentum: $g_{i+1} = \mu \cdot g_i + \frac{\nabla_{\delta} \mathcal{L}}{||\nabla_{\delta} \mathcal{L}||_1}$
			\STATE Update Perturbation: $\delta_{i+1} = \text{Clip}_{\epsilon}(\delta_i - \alpha \cdot \text{sign}(g_{i+1}))$
			\ENDFOR
			\RETURN $\tilde{S}_t^{(N)} = S_t(v_{target}+\delta_N)$
		\end{algorithmic}
	\end{algorithm}
	
	\section{Experiments}
	
	\subsection{Experimental Setup}
	\begin{itemize}[itemsep=0pt,topsep=2pt]
		\item \textbf{Dataset (E-ShopBench):} 200 adversarial scenarios across Amazon (80), eBay (60), and Taobao (60). Each scenario contains one target (expensive) and one distractor (cheap) item.
		\item \textbf{Victim Models:} \textit{white-box:} Mobile-Agent-v2 (Qwen-VL-Chat), AppAgent (LLaVA-1.6-Vicuna-7B). \textit{black-box:} GPT-4o, Gemini-1.5-Pro, Claude-3.5-Sonnet.
		\item \textbf{Baselines:} clean, Random Noise ($\epsilon=8/255$), Typographic (``Best Deal'' overlay), Adv. Pop-up \citep{zhang2025attacking}.
		\item \textbf{Metrics:} ASR (Attack Success Rate), LPIPS (perceptual similarity).
		\item \textbf{Statistical Reporting:} Scenario-level ASR numbers are reported as descriptive percentages; tables use rounded values for readability, and the prose emphasizes effect sizes and broad trends. We reserve mean $\pm$ std style reporting for distributional summaries such as attention weights.
	\end{itemize}

	\paragraph{Metrics and Statistical Reporting.}
	ASR is defined as the percentage of evaluation scenarios in which the agent selects the pre-defined expensive target item (scenario-level). Unless otherwise specified, scenario-level ASR values should be read as descriptive summaries over E-ShopBench rather than formal inferential estimates. Results reported with $\pm$ in this paper denote mean $\pm$ std for underlying continuous or token-level measurements, not 95\% confidence intervals.

	\subsection{Main Results}
	Table \ref{tab:main} reports white-box ASR on E-ShopBench.
	
	\begin{table}[t]
		\centering
		\caption{White-box attack performance on E-ShopBench ($\epsilon=8/255$). Values are descriptive scenario-level ASR percentages, rounded for readability.}
		\label{tab:main}
		\resizebox{\columnwidth}{!}{%
			\begin{tabular}{llc}
				\toprule
				\textbf{Method} & \textbf{Victim Agent} & \textbf{ASR (\%)}$\uparrow$ \\
				\midrule
				clean & Mobile-Agent-v2 & 4 \\
				Random Noise & Mobile-Agent-v2 & 7 \\
				Typographic & Mobile-Agent-v2 & 32 \\
				Adv. Pop-up & Mobile-Agent-v2 & 45 \\
				\textbf{PriceBlind} & \textbf{Mobile-Agent-v2} & \textbf{82} \\
				\midrule
				clean & AppAgent & 4 \\
				\textbf{PriceBlind} & \textbf{AppAgent} & \textbf{79} \\
				\bottomrule
			\end{tabular}%
		}
	\end{table}

	PriceBlind substantially degrades price-constrained decision-making in white-box settings. On Mobile-Agent-v2, PriceBlind reaches $82\%$ ASR, compared with $32\%$ for Typographic and $45\%$ for Adv. Pop-up. On AppAgent, PriceBlind reaches $79\%$ while the clean condition remains at $4\%$. These are descriptive scenario-level results on the main E-ShopBench screenshot-based benchmark and serve as the paper's central white-box empirical claim.

	\subsection{Black-box Transferability Results}
	Table \ref{tab:blackbox} presents results on black-box proprietary models.
	
	\begin{table}[t]
		\centering
		\caption{Black-box transfer ASR (\%) on proprietary models under the single-turn coordinate-selection protocol (E-ShopBench, $n=200$ scenarios). Values are descriptive and rounded for readability.}
		\label{tab:blackbox}
		\resizebox{\columnwidth}{!}{%
			\begin{tabular}{lccc}
				\toprule
				\textbf{Method} & \textbf{GPT-4o} & \textbf{Gemini-1.5-Pro} & \textbf{Claude-3.5} \\
				\midrule
				clean & 2 & 2 & 2 \\
				Typographic & 19 & 15 & 13 \\
				Adv. Pop-up & 22 & 20 & 18 \\
				\midrule
				PriceBlind (Single) & 13 & 11 & 9 \\
				PriceBlind (Ensemble) & \textbf{41} & \textbf{39} & \textbf{35} \\
				\bottomrule
			\end{tabular}%
		}
	\end{table}

	Under the single-turn coordinate-selection protocol, the ensemble strategy substantially improves black-box transfer. In Table~\ref{tab:blackbox}, ``PriceBlind (Single)'' denotes the single-surrogate Qwen-VL setting. On GPT-4o, ASR rises from $13\%$ for this single-surrogate setup to $41\%$ for the ensemble; on Gemini-1.5-Pro, from $11\%$ to $39\%$; and on Claude-3.5, from $9\%$ to $35\%$. These black-box numbers should be interpreted as simplified protocol results rather than end-to-end multi-turn agent execution. Together with Table~\ref{tab:main}, they define the paper's main benchmark evidence; the appendix prompt-style stress tests are supplementary and are not intended as directly comparable replacements.
	
	\subsection{Qualitative Analysis}
	As a supplementary, non-systematic qualitative analysis, we manually inspected Chain-of-Thought (CoT) logs from successful attacks and grouped the observed rationales into three descriptive failure patterns. These language-level explanations are post-hoc rationalizations rather than direct observations of the optimized embedding direction, so the patterns below should be interpreted as symptoms of disrupted price-value association rather than standalone proof of mechanism. The percentages below denote descriptive shares within the inspected successful attacks, not benchmark-level ASR values:
	
	\textbf{Pattern 1: Visual-Semantic Overriding} (45\%): The agent acknowledges the high price but still reframes the item as compatible with a value-oriented choice. For instance: "Although this item is \$499, it looks like the basic option or a better-value package."
	
	\textbf{Pattern 2: Numerical Blindness} (30\%): The agent simply ignored the text. When asked to "buy the cheapest", the agent clicked the expensive perturbed item stating: "This looks like the most basic model."
	
	\textbf{Pattern 3: Layout Hallucination} (25\%): The perturbation caused the agent to misalign the price text, believing a neighbor item's price belonged to the target item.

	Taken together, these patterns suggest that the perturbation disrupts price-value association, but they should not be read as evidence that the representation literally becomes a cheap-item embedding. The same visual shift may be rationalized downstream as a basic model, a discounted/bundle interpretation, or a price-layout mismatch.
	
	\begin{figure}[t]
		\centering
		\includegraphics[width=\columnwidth]{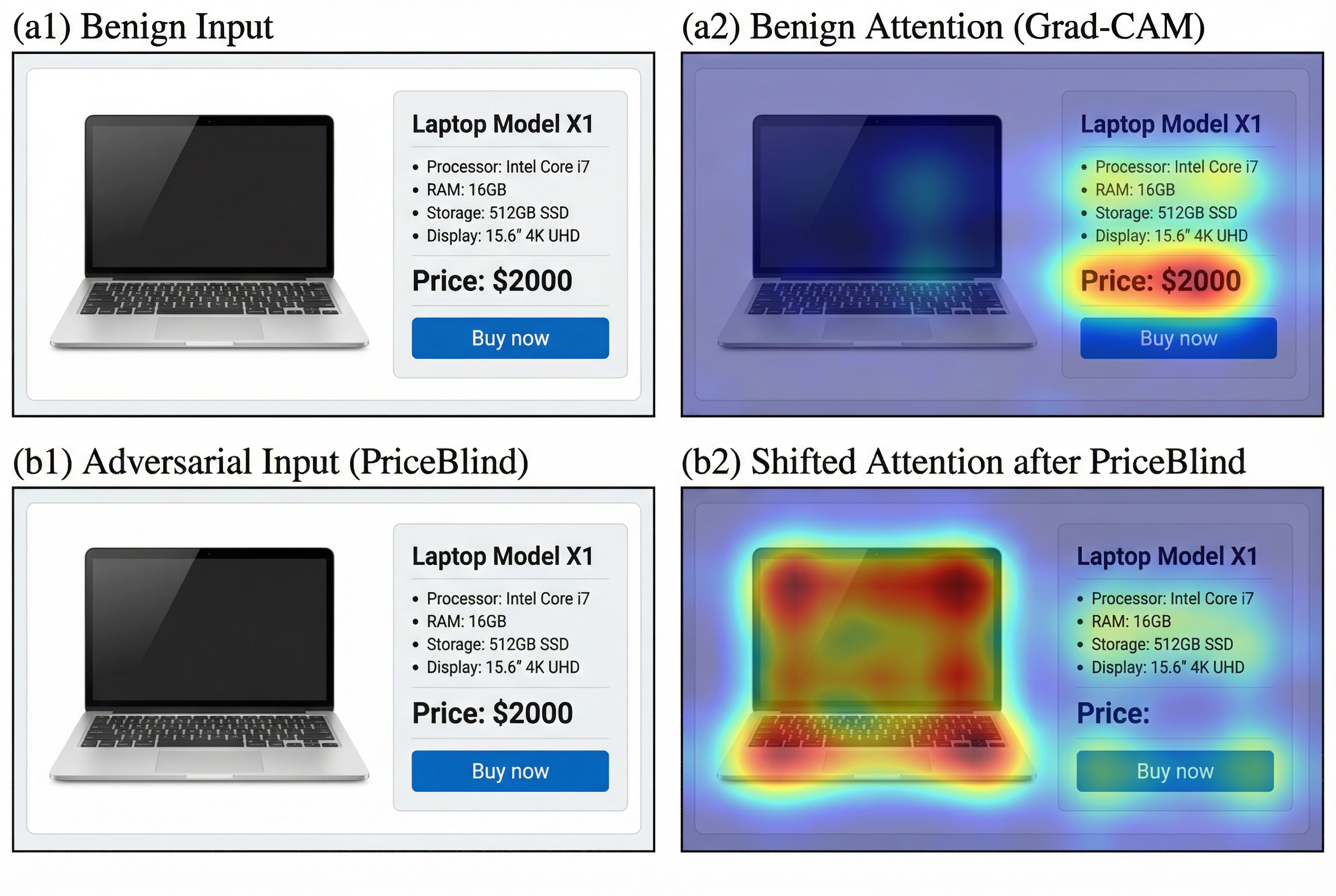}
		\caption{Grad-CAM saliency visualization. In the benign case (Top), the model highlights the OCR price regions. Under PriceBlind attack (Bottom), saliency shifts away from the price text towards the manipulated visual features.}
		\label{fig:attention}
	\end{figure}
	
	\subsection{Attention Weight Analysis}
	Figure~\ref{fig:attention} provides a qualitative Grad-CAM saliency visualization, while the table below reports supplementary quantitative evidence from Qwen-VL's cross-attention layers. Together, they provide descriptive support for the proposed mechanism rather than a second benchmark.
	
	\begin{table}[t]
		\centering
		\caption{Attention weight distribution (mean $\pm$ std) across token types.}
		\label{tab:attention}
		\resizebox{\columnwidth}{!}{%
			\begin{tabular}{lccc}
				\toprule
				\textbf{Condition} & \textbf{Visual} & \textbf{Price Text} & \textbf{Other} \\
				\midrule
				clean & $0.32_{\pm 0.05}$ & $0.28_{\pm 0.04}$ & $0.40_{\pm 0.06}$ \\
				PriceBlind & $0.58_{\pm 0.07}$ & $0.11_{\pm 0.03}$ & $0.31_{\pm 0.05}$ \\
				\midrule
				\textbf{Change} & \textbf{+81\%} & \textbf{-61\%} & -22\% \\
				\bottomrule
			\end{tabular}%
		}
	\end{table}

	\textbf{Supplementary finding:} PriceBlind shifts cross-attention mass from price text tokens to visual tokens. \textbf{Numeric evidence:} Visual attention increases from $0.32_{\pm 0.05}$ to $0.58_{\pm 0.07}$ (+81\%), while price-text attention decreases from $0.28_{\pm 0.04}$ to $0.11_{\pm 0.03}$ (-61\%); ``Other'' tokens drop by 22\%. \textbf{Scope caveat:} These token-level summaries are descriptive statistics over analyzed trajectories and provide supporting evidence consistent with the mechanism, rather than a standalone causal proof.

	\subsection{Ablation Study}
	
	\begin{table}[t]
		\centering
		\caption{Ablation on Mobile-Agent-v2 (white-box). ASR values are descriptive and rounded for readability; LPIPS is lower-is-better perceptual distance.}
		\label{tab:ablation}
		\resizebox{\columnwidth}{!}{%
		\begin{tabular}{lcc}
			\toprule
			\textbf{Configuration} & \textbf{ASR (\%)} & \textbf{LPIPS}$\downarrow$ \\
			\midrule
			$\mathcal{L}_{action}$ only & 45 & $0.028$ \\
			$\mathcal{L}_{dec}$ only & 39 & $0.031$ \\
			Full ($\lambda=1.5$) & \textbf{82} & $0.033$ \\
			\midrule
			w/o Momentum & 69 & $0.031$ \\
			w/o DI Transform & 71 & $0.032$ \\
			Single Model (Qwen) & 76 & $0.031$ \\
			\bottomrule
		\end{tabular}}
	\end{table}
	
	Both $\mathcal{L}_{action}$ and $\mathcal{L}_{dec}$ are necessary, and ensemble training is important for black-box transfer under the single-turn coordinate-selection protocol. On Mobile-Agent-v2, either loss alone remains below $50\%$ ASR, while the full method reaches about $80\%$; separately, transfer to GPT-4o rises from the low teens for a single surrogate to about $40\%$ for the ensemble. The ablation ASR/LPIPS values are descriptive white-box results on Mobile-Agent-v2, while the GPT-4o transferability numbers come from the simplified single-turn protocol.
	
	\begin{figure}[t]
		\centering
		\includegraphics[width=\columnwidth]{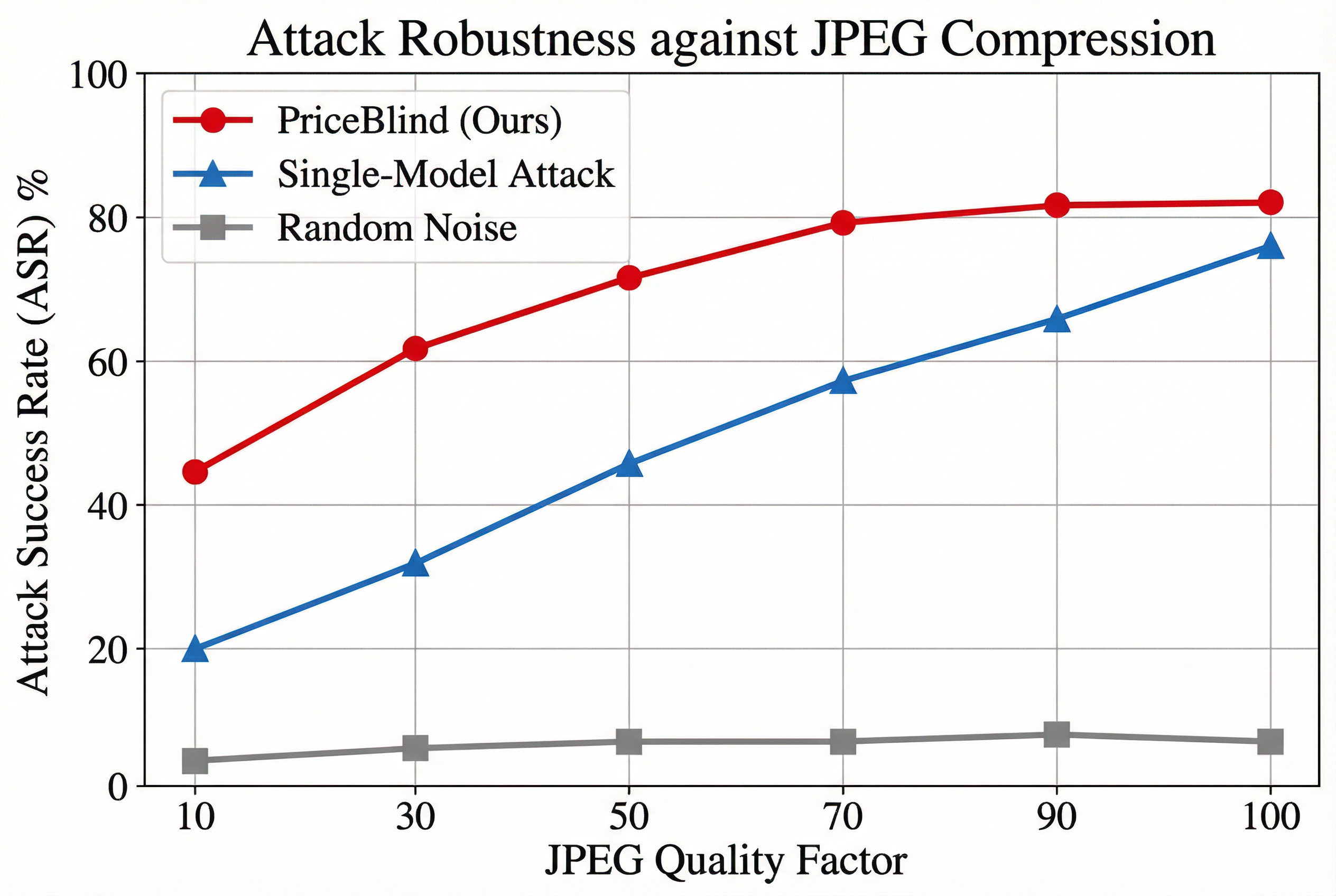}
		\caption{JPEG robustness curves. PriceBlind maintains high ASR even under aggressive JPEG compression and consistently outperforms baselines across the tested JPEG settings.}
		\label{fig:robustness}
	\end{figure}
	
	\subsection{Defense Evaluation}
	\label{sec:defense}
	
	\begin{table}[t]
		\centering
		\caption{Defense results on Mobile-Agent-v2 in the screenshot-based setting. ASR and clean accuracy are descriptive percentages rounded for readability.}
		\label{tab:defense}
		\resizebox{\columnwidth}{!}{%
		\begin{tabular}{lcc}
			\toprule
			\textbf{Defense} & \textbf{PriceBlind ASR} & \textbf{Clean Acc.} \\
			\midrule
			None (clean reference) & 82 & 96 \\
			\midrule
			JPEG Compression (Q=50) & 72 & 94 \\
			Gaussian Blur ($\sigma$=1.0) & 68 & 92 \\
			\midrule
			Robust-CLIP \citep{schlarmann2024robust} & 58 & 93 \\
			AdPO \citep{liu2025adpo} & 53 & 95 \\
			\midrule
			Verify-then-Act (Hard) & 12 & 88 \\
			VtA + Robust-CLIP & \textbf{9} & 87 \\
			\bottomrule
		\end{tabular}}
	\end{table}
	
	Within the main screenshot-based Mobile-Agent-v2 benchmark, robust encoders partially mitigate PriceBlind, while Verify-then-Act provides the strongest reduction. ASR drops from about $82\%$ with no defense to about $58\%$/$53\%$ under Robust-CLIP and AdPO, and to the low teens under VtA; combining VtA with Robust-CLIP reduces ASR to below $10\%$. This stronger verification-heavy stack comes with a clean-accuracy trade-off, which falls into the high-$80\%$ range.
	
	\paragraph{Why Robust Encoders Provide Partial Defense.}
	We hypothesize that robust encoders primarily defend against perturbations that cause large, arbitrary embedding shifts. However, PriceBlind's semantic decoupling loss creates perturbations that move embeddings \textit{within} the natural image manifold (towards a low-cost/value-associated anchor region), which robust training does not specifically address. This suggests that defending against semantic manipulation attacks may require different approaches than defending against random perturbations---specifically, methods that preserve semantic consistency between visual features and associated metadata.
	
	\subsection{Comparison with Related Attack Methods}
	We compare PriceBlind with recent transfer attack methods in Table~\ref{tab:comparison_main}. To ensure fair comparison, we re-implemented baseline attacks under controlled conditions on E-ShopBench, following the spirit of reliable robustness evaluation \citep{croce2020reliable}.
	
	\begin{table}[t]
		\centering
		\caption{Controlled attack comparison on E-ShopBench ($\epsilon=8/255$). Transfer columns use the single-turn coordinate-selection protocol; values are descriptive ASR (\%) rounded for readability.}
		\label{tab:comparison_main}
		\resizebox{\columnwidth}{!}{%
		\begin{tabular}{lcccc}
			\toprule
			\textbf{Method} & \textbf{white-box} & \textbf{GPT-4o} & \textbf{Gemini} & \textbf{Stealth} \\
			\midrule
			MI-FGSM & 65 & 18 & 16 & High \\
			DI-FGSM & 69 & 22 & 20 & High \\
			X-Transfer & 76 & 36 & 32 & High \\
			AdvDiffVLM & 73 & 33 & 30 & High \\
			Pop-up Attack & 45 & 22 & 20 & Low \\
			\textbf{PriceBlind} & \textbf{82} & \textbf{41} & \textbf{39} & \textbf{High} \\
			\bottomrule
		\end{tabular}}
	\end{table}
	
	PriceBlind outperforms all controlled baseline attacks on both white-box and simplified transfer columns. White-box ASR is about $82\%$ versus a strongest baseline in the mid-$70\%$ range, and GPT-4o/Gemini transfer reaches about $41\%$/$39\%$ versus strongest baselines in the mid-$30\%$ and low-$30\%$ ranges. Transfer columns are measured under a single-turn coordinate-selection protocol and should not be interpreted as end-to-end multi-turn task completion rates.
	
	\subsection{Preliminary Probe to Alternative Encoders}
	As a limited supplementary probe beyond our main screenshot-based evaluation, we examined transferability to several alternative encoder architectures:
	
	\begin{table}[t]
		\centering
		\caption{Preliminary transferability probe to alternative encoder architectures. Values are descriptive ASR (\%) rounded for readability.}
		\label{tab:nonclip_main}
		\resizebox{\columnwidth}{!}{%
		\begin{tabular}{lcc}
			\toprule
			\textbf{Target Encoder} & \textbf{ASR (\%)} & \textbf{$\Delta$ vs CLIP} \\
			\midrule
			CLIP-ViT-L/14 (reference) & 82 & -- \\
			SigLIP-SO400M & 45 & $-37$ \\
			EVA-CLIP-8B & 52 & $-30$ \\
			InternViT-6B & 39 & $-43$ \\
			\bottomrule
		\end{tabular}}
	\end{table}
	
	In this preliminary probe, transferability decreases on alternative encoders but remains non-trivial. Relative to the CLIP reference, EVA-CLIP stays above $50\%$, while SigLIP and InternViT drop to roughly the mid-$40\%$ and high-$30\%$ range, respectively. These values should be read as indicative supplementary checks rather than definitive cross-encoder estimates, and not as replacements for the main screenshot-based benchmark.
	
	\subsection{Layer-wise Attention Analysis}
	As a supplementary mechanistic probe, we analyzed attention weights across all 32 transformer layers of Qwen-VL. We found that the attention shift is most pronounced in layers 20-28 (the ``reasoning'' layers), with Pearson correlation $r=0.73$ ($p<0.001$, $n=32$ layers) between layer depth and attention shift magnitude. Early layers (1-10) show minimal change, suggesting that low-level visual features remain intact while high-level semantic interpretation is manipulated. This layer-wise analysis is intended as supporting evidence rather than part of the paper's central benchmark claim.
	
	As supplementary, non-systematic checks beyond our main screenshot-focused evaluation, Appendix~\ref{sec:appendix_structured} reports preliminary results for DOM/AX-tree observation settings, and Appendix~\ref{sec:appendix_instruction} summarizes prompt-style stress tests that are not intended as directly comparable replacements for the main benchmark.

	\section{Discussion}
	
	\paragraph{Root Cause: Dataset Bias in Pre-training.}
	We argue that the root cause of VDH is not just the architecture, but the pre-training data. Datasets like LAION-5B contain massive amounts of noisy image-text pairs where the model learns that "visual appearance" implies "textual attributes". PriceBlind exploits this correlation by nudging the embedding toward low-cost/value-associated directions, effectively leveraging the model's own statistical biases against it.
	
	\paragraph{Novelty Relative to CLIP-Space Attacks.}
	We acknowledge that embedding-space manipulation for VLMs has been explored in prior work. PriceBlind's contribution is: (1) the identification of VDH as a specific vulnerability pattern in agentic financial tasks, (2) the semantic decoupling loss that targets price-value associations rather than generic class boundaries, and (3) a broad screenshot-based evaluation on E-ShopBench showing strong vulnerability in the main clear-constraint benchmark and non-trivial transfer under our evaluated single-turn black-box protocol.
	
	\paragraph{The Arms Race: Attack vs. Defense.}
	Our work highlights a classic security arms race. While we propose PriceBlind, defenses like separate OCR modules can mitigate it. However, attackers can counter this by attacking the OCR engine itself. This suggests that a static defense is insufficient. Future agent architectures must incorporate \textit{dynamic verification}, where the agent actively seeks corroborating evidence when visual and textual signals conflict.
	
	\paragraph{Implications for Future Agents.}
	The vulnerability exposed by PriceBlind suggests that current "Late Fusion" architectures are insufficient for high-stakes decision making. Future agents may need to adopt a "Verify-then-Act" architecture, where critical constraints are verified by specialized, non-neural symbolic modules before the neural planner is allowed to execute an action. Reliance on a single, monolithic MLLM for both perception and reasoning is a single point of failure.
	
	\paragraph{Broader Impact on Autonomous Systems.}
	Beyond e-commerce, the Visual Dominance Hallucination vulnerability has significant implications for autonomous systems in healthcare, finance, and critical infrastructure. Medical imaging agents could be manipulated to misinterpret diagnostic images, while financial document processing systems might be deceived about numerical values in contracts or invoices. Our findings suggest that any system where visual perception directly influences high-stakes decisions requires additional verification layers. The fundamental tension between efficiency (single-model inference) and security (multi-modal verification) will shape the design of next-generation autonomous agents.
	
	\paragraph{Recommendations for Practitioners.}
	Based on our findings, we recommend the following for deploying vision-language agents in production: (1) implement redundant verification for price-critical decisions using separate OCR pipelines, (2) establish confidence thresholds that trigger human review when visual-textual conflicts are detected, (3) maintain audit logs of agent decisions for post-hoc analysis, and (4) consider ensemble approaches that aggregate predictions from multiple encoder architectures to reduce single-point vulnerabilities.
	
	\section{Conclusion}
	We presented \textbf{PriceBlind}, a controlled study of screenshot-based visual vulnerability in e-commerce agents. On our main screenshot-based E-ShopBench benchmark with clear price constraints, we provided evidence that ``Visual Dominance'' is an important weakness in current MLLMs, allowing attackers to override price evidence via stealthy image perturbations in white-box settings. Under the evaluated single-turn coordinate-selection protocol in a simplified layout-aware setting, black-box transfer remains non-trivial, while stronger verification-heavy stacks can substantially reduce risk at some clean-accuracy cost. As agents become autonomous economic actors, solving this \textbf{Instruction-Perception Conflict} is a prerequisite for robust deployment.
	
	Our key findings can be summarized as follows: (1) On our main screenshot-based E-ShopBench benchmark with clear price constraints, Visual Dominance Hallucination yields white-box ASR around 80\%; (2) The semantic decoupling loss improves attack effectiveness by disrupting price-value associations in the embedding space; (3) Black-box transfer remains non-trivial (roughly 35--41\% ASR) under the evaluated single-turn coordinate-selection protocol; (4) Standard robust encoders reduce ASR only partially, whereas Verify-then-Act combined with robust encoders can reduce ASR to below 10\% with clean accuracy in the high-$80\%$ range.
	
	Looking forward, we believe that the security of autonomous agents will become increasingly critical as these systems handle more financial transactions. The community must develop principled approaches to verify agent decisions, particularly when visual and textual signals conflict. We hope that PriceBlind serves as a wake-up call for the development of more robust multimodal architectures.
	
	\section*{Limitations}
	\textbf{Benchmark Scale.} E-ShopBench contains 200 scenarios, which is modest compared to large-scale benchmarks. Future work should evaluate on larger benchmarks with more diverse UI templates.
	
	\textbf{CLIP Encoder Dependency.} Our attack is optimized for CLIP-based visual encoders. Our preliminary cross-encoder probe suggests lower transferability on alternative encoders overall, especially on the true non-CLIP encoders SigLIP \citep{zhai2023sigmoid} and InternViT, but these supplementary checks are not yet a definitive cross-encoder benchmark.
	
	\textbf{Layout-Aware Optimization.} Our formal threat model already assumes screenshot-conditioned or stable-layout target coordinates during perturbation generation. Broader dynamic UI changes may therefore reduce attack success outside this controlled setting.
	
	\textbf{Computational Cost.} Per-image perturbation requires $\sim$45 seconds on a single A100 GPU. We did not explore Universal Adversarial Perturbations which could amortize the cost.
	
	\textbf{Real-World Deployment Gap.} Our evaluation uses simulated e-commerce environments rather than live production systems. Real-world platforms may employ additional security measures, rate limiting, or anomaly detection that could affect attack success rates. The gap between controlled experiments and real-world deployment remains an important consideration for interpreting our results.
	
	\textbf{User Intent Variability.} Our main evaluation protocol prioritizes clear price constraints to isolate instruction-following failures. We additionally report implicit and vague prompt settings as supplemental stress tests (Appendix~\ref{sec:appendix_instruction}), but those checks are intended to illustrate trend direction rather than serve as directly comparable replacements for the main benchmark figures. Broader real-world intent diversity remains under-explored and may affect observed attack rates.
	
	Additional limitations including black-box evaluation methodology, theoretical analysis scope, and screenshot-centric evaluation scope are discussed in Appendix~\ref{sec:appendix_limitations}.
	
	\section*{Ethics Statement}
	All experiments were conducted in a restricted sandbox environment. No real transactions were executed, and no adversarial images were uploaded to live public e-commerce platforms. We believe that responsible disclosure of these vulnerabilities is essential for improving the security of autonomous agents before they are widely deployed in high-stakes financial applications.

\paragraph{Responsible Disclosure.} We encourage the research community to use our findings constructively to develop more robust multimodal architectures rather than for malicious purposes.
	
	\paragraph{Broader Implications.} The proliferation of autonomous agents in financial transactions raises fundamental questions about accountability and trust. Our work highlights the need for regulatory frameworks governing autonomous economic actors. We hope our findings contribute to the development of safer AI systems.

	\section*{Acknowledgements}
	Generative AI tools were used to assist with language polishing and to draft parts of the experimental code. All scientific content, code, and results were reviewed, edited, and validated by the authors, who take full responsibility for the work.
	
	\bibliography{custom}
	
	\appendix
	
	\section{Extended Heuristic Analysis}
	\label{sec:appendix_theory}
	
	\subsection{Cross-Attention Dynamics}
	Let $\mathbf{V} = [v_1, ..., v_n]$ be the visual token sequence and $\mathbf{T} = [t_1, ..., t_m]$ be the textual token sequence. In the cross-attention layer:
	\begin{equation}
		\text{Attn}(q, \mathbf{K}, \mathbf{V}) = \text{softmax}\left(\frac{q\mathbf{K}^T}{\sqrt{d}}\right)\mathbf{V}
	\end{equation}
	where $\mathbf{K} = [\mathbf{K}_v; \mathbf{K}_t]$ concatenates visual and textual keys.
	
	\textit{Heuristic derivation for Proposition 1.} Let
	\begin{equation}
		\begin{aligned}
			w_v^{(0)} &= \frac{\exp(q \cdot k_v^{(0)} / \sqrt{d})}{Z^{(0)}}, \\
			w_v^{(1)} &= \frac{\exp(q \cdot (k_v^{(0)} + \Delta_v) / \sqrt{d})}{Z^{(1)}}.
		\end{aligned}
	\end{equation}
	Under the approximation $Z^{(1)} \approx Z^{(0)}$, we obtain
	\begin{equation}
		\frac{w_v^{(1)}}{w_v^{(0)}} \approx \exp\left(\frac{q \cdot \Delta_v}{\sqrt{d}}\right).
	\end{equation}
	If the relevant textual logits are approximately unchanged, then $w_t^{(1)} \approx w_t^{(0)}$, which gives
	\begin{equation}
		\frac{w_v^{(1)}}{w_t^{(1)}} \approx \frac{w_v^{(0)}}{w_t^{(0)}} \exp\left(\frac{q \cdot \Delta_v}{\sqrt{d}}\right).
	\end{equation}
	Therefore, $w_v^{(1)} > w_t^{(1)}$ is implied when $q \cdot \Delta_v > \sqrt{d}\log(w_t^{(0)} / w_v^{(0)})$. This argument is heuristic: in real MLLMs, multi-layer fusion, residual connections, and normalization can all affect the exact threshold.
	
	\textbf{Limitations of This Analysis.} This analysis assumes a simplified single-layer cross-attention model. Real MLLMs employ multiple attention layers with residual connections, layer normalization, and varying fusion strategies. The exact conditions for successful attack may vary across architectures.
	
\subsection{Connection to CLIP Modality Gap}
The effectiveness of $\mathcal{L}_{dec}$ can be explained through the geometry of CLIP's latent space. Studies have shown that CLIP embeddings exhibit a "modality gap" where image and text embeddings occupy different regions of the hypersphere \citep{liang2022mind}. Within the image manifold, semantic concepts form distinct clusters. Our attack exploits this structure by projecting the perturbed target-item embedding toward a low-cost/value-associated anchor region while maintaining pixel-level similarity.
	
Let $\mathcal{M}_{cheap}$ denote the low-cost anchor manifold and $\mathcal{M}_{luxury}$ the original luxury-item neighborhood. The semantic decoupling loss effectively solves:
\begin{equation}
		\min_{\delta} \, d(\mathcal{E}_v(v_{target} + \delta), \mathcal{M}_{cheap}) \quad \text{s.t.} \quad ||\delta||_\infty \leq \epsilon
\end{equation}
	
	This geometric perspective explains why the attack transfers across models: all CLIP-based VLMs share similar embedding space geometry.
	
\subsection{Momentum Smoothing Intuition}
Momentum acts as an exponential moving average over successive gradients, which can damp abrupt direction changes even though it does not imply a literal reduction in the raw variance of the accumulated state. A schematic form of the update is:
\begin{equation}
		g_{i+1} \approx \mu \cdot g_i + \nabla_{\delta}\mathcal{L}(\delta_i)
\end{equation}
Unrolling this recurrence shows that the current update direction aggregates information from multiple previous steps with exponentially decaying weights. In our setting, this temporal smoothing helps stabilize optimization trajectories under the diverse-input transform and is best interpreted as a heuristic transferability aid rather than a formal variance-reduction guarantee.
	
	\section{Extended Ablation Studies}
	\label{sec:appendix_ablation}
	
	\subsection{Sensitivity to Lambda}
	
	\begin{table}[h]
		\centering
		\caption{Extended ablation on the decoupling weight (lambda). ASR values are descriptive and rounded for readability.}
		\label{tab:lambda_ablation}
		\begin{tabular}{lcc}
			\toprule
			\textbf{Configuration} & \textbf{ASR (\%)} & \textbf{LPIPS}$\downarrow$ \\
			\midrule
			Full ($\lambda=0.5$) & 72 & $0.030$ \\
			Full ($\lambda=1.0$) & 79 & $0.032$ \\
			Full ($\lambda=1.5$) & \textbf{82} & $0.033$ \\
			Full ($\lambda=2.0$) & 82 & $0.038$ \\
			Full ($\lambda=3.0$) & 79 & $0.052$ \\
			\bottomrule
		\end{tabular}
	\end{table}
	
	\textbf{Key Findings:}
	\begin{itemize}
		\item \textbf{Lower-$\lambda$ regime (e.g., $\lambda=0.5$):} The attack behaves more like a standard coordinate attack. ASR is lower because the agent more often reads the price and aborts.
		\item \textbf{Mid-$\lambda$ regime (1.0 - 2.0):} The ASR peaks. The image features are successfully decoupled from the "expensive" concept.
		\item \textbf{At $\lambda=3.0$:} Visual quality degrades substantially (LPIPS = 0.052 $>$ 0.05), indicating the imperceptibility constraint is no longer well preserved at this setting.
	\end{itemize}
	
	\subsection{Component Contribution Analysis}
	
	\begin{table}[h]
		\centering
		\caption{Component contribution analysis based on descriptive scenario-level ASR values.}
		\label{tab:component}
        \resizebox{\linewidth}{!}{
		\begin{tabular}{lccc}
			\toprule
			\textbf{Configuration} & \textbf{ASR} & \textbf{$\Delta$ vs clean} & \textbf{Contribution} \\
			\midrule
			clean & $4\%$ & -- & -- \\
			$\mathcal{L}_{action}$ only & $45\%$ & $+41$ pp & $52\%$ \\
			$\mathcal{L}_{dec}$ only & $39\%$ & $+35$ pp & $45\%$ \\
			Full PriceBlind & $82\%$ & $+78$ pp & $100\%$ \\
			\bottomrule
		\end{tabular}}
	\end{table}
	
	Using a unified Mobile-Agent-v2 clean reference (clean ASR $\approx 4\%$), coordinate targeting ($\mathcal{L}_{action}$) accounts for roughly half of the total improvement, while semantic decoupling ($\mathcal{L}_{dec}$) contributes slightly under half. The remaining few percentage points correspond to interaction/synergy between the two components.
	
	\subsection{Surrogate Model Sensitivity}
	
	\begin{table}[h]
		\centering
		\caption{Surrogate model ablation for GPT-4o transfer. Values are descriptive and rounded for readability.}
		\label{tab:surrogate}
        \resizebox{\linewidth}{!}{
		\begin{tabular}{lc}
			\toprule
			\textbf{Surrogate Configuration} & \textbf{GPT-4o ASR (\%)} \\
			\midrule
			Single Model (Qwen-VL) & 13 \\
			Single Model (LLaVA-1.6) & 16 \\
			Ensemble (Qwen + LLaVA) & 41 \\
			Ensemble (+ MiniGPT-4) & 43 \\
			Ensemble (+ InternVL) & 45 \\
			\bottomrule
		\end{tabular}}
	\end{table}
	
	This descriptive ablation suggests that attacking the common intersection of CLIP-based models is sufficient to obtain most of the downstream transfer, with diminishing returns after two strong surrogates.
	
	\section{Transferability to Alternative Encoders}
	\label{sec:appendix_nonclip}
	
	\begin{table}[h]
		\centering
		\caption{Preliminary transferability probe to alternative encoder architectures. Values are descriptive and rounded for readability.}
		\label{tab:nonclip}
        \resizebox{\linewidth}{!}{
		\begin{tabular}{lcc}
			\toprule
			\textbf{Target Encoder} & \textbf{ASR (\%)} & \textbf{$\Delta$ vs CLIP} \\
			\midrule
			CLIP-ViT-L/14 (reference) & 82 & -- \\
			SigLIP-SO400M & 45 & $-37$ \\
			EVA-CLIP-8B & 52 & $-30$ \\
			InternViT-6B & 39 & $-43$ \\
			\bottomrule
		\end{tabular}}
	\end{table}
	
	This preliminary probe suggests that the attack partially exploits CLIP-specific embedding geometry. EVA-CLIP remains above $50\%$ ASR, while the true non-CLIP encoders SigLIP and InternViT drop to roughly the mid-$40\%$ and high-$30\%$ range. We therefore treat these results as indicative supplementary evidence rather than a definitive cross-encoder study.
	
	\section{Comparison with Related Attack Methods}
	\label{sec:appendix_comparison}
	
	\begin{table}[h]
		\centering
		\caption{Controlled comparison with related attack methods on E-ShopBench. Values are descriptive ASR percentages rounded for readability.}
		\label{tab:comparison}
		\resizebox{\columnwidth}{!}{%
		\begin{tabular}{lcccc}
			\toprule
			\textbf{Method} & \textbf{white-box} & \textbf{GPT-4o} & \textbf{Gemini} & \textbf{Stealth} \\
			\midrule
			MI-FGSM \citep{dong2018boosting} & 65 & 18 & 16 & High \\
			DI-FGSM & 69 & 22 & 20 & High \\
			SSA \citep{naseer2020ssa} & 71 & 29 & 25 & High \\
			X-Transfer \citep{huang2025xtransfer} & 76 & 36 & 32 & High \\
			AdvDiffVLM \citep{guo2024efficient} & 73 & 33 & 30 & High \\
			Pop-up Attack & 45 & 22 & 20 & Low \\
			\textbf{PriceBlind} & \textbf{82} & \textbf{41} & \textbf{39} & \textbf{High} \\
			\bottomrule
		\end{tabular}}
	\end{table}
	
	PriceBlind outperforms all baselines. The key advantage comes from our semantic decoupling loss, which specifically targets the price-value association rather than generic classification boundaries.
	
	\section{Structured Observation Agents}
	\label{sec:appendix_structured}
	
	\paragraph{DOM/AX-Tree Agents.} Agents that operate on DOM or accessibility tree representations receive structured element attributes including text content, element type, and bounding boxes. Since price information is typically encoded as text attributes in the DOM, these agents may be more robust to visual perturbations.
	
	\begin{table}[h]
		\centering
		\caption{Preliminary descriptive comparison with grounding-targeted attacks across observation modalities.}
		\label{tab:grounding}
        \resizebox{\linewidth}{!}{
		\begin{tabular}{lccc}
			\toprule
			\textbf{Attack Type} & \textbf{Screenshot} & \textbf{DOM+Screenshot} & \textbf{DOM-only} \\
			\midrule
			Coordinate-only & $45\%$ & $28\%$ & $8\%$ \\
			PriceBlind (Full) & $82\%$ & $52\%$ & $13\%$ \\
			\midrule
			\textbf{Gain from $\mathcal{L}_{dec}$} & $+37$ pp & $+24$ pp & $+5$ pp \\
			\bottomrule
		\end{tabular}}
	\end{table}
	
	These preliminary descriptive results suggest that semantic decoupling provides clear gains in screenshot-only and DOM+Screenshot settings. In DOM-only settings, the residual gain is about $+5$ percentage points and should not be over-interpreted as direct evidence for the same visual mechanism. For screenshot-only agents, $\mathcal{L}_{dec}$ contributes roughly $+37$ percentage points. These supplementary modality comparisons are not intended to replace the main screenshot benchmark.
	
	\section{Extended Limitations}
	\label{sec:appendix_limitations}
	
	\paragraph{Black-box API Evaluation Methodology.} Our black-box evaluation on GPT-4o, Gemini-1.5-Pro, and Claude-3.5-Sonnet uses a single-turn coordinate selection setup where models receive the screenshot and instruction, then output a click coordinate. This design choice isolates the visual perception vulnerability and enables reproducible evaluation, but differs from full multi-turn agent deployments that include additional reasoning steps, tool use, and error recovery. The reported ASR quantifies vulnerability under this simplified protocol and does not directly estimate production multi-turn vulnerability.
	
	\paragraph{Theoretical Analysis Scope.} The cross-attention analysis provides intuition for the attention hijacking mechanism but does not tightly characterize the exact conditions under which the attack succeeds across all possible fusion architectures (early fusion, late fusion, cross-attention variants). A more rigorous analysis connecting specific architectural choices to vulnerability would strengthen the theoretical contribution.
	
	\paragraph{White-box Surrogate Requirement.} PriceBlind requires white-box access to at least one surrogate model. Fully black-box optimization (query-based) would be more practical but computationally expensive.
	
	\paragraph{Screenshot-Centric Evaluation.} Our main evaluation focuses on screenshot-based agents. We include a preliminary, non-systematic analysis of DOM/AX-tree observation settings in Appendix~\ref{sec:appendix_structured}, but this does not yet constitute a comprehensive study across modern structured-observation agent pipelines. Recent benchmarks like OmniACT \citep{kapoor2024omniact} and agent frameworks leveraging Set-of-Mark prompting \citep{yang2024setofmark} remain important targets for future work.
	
	\section{Detailed Experimental Configuration}
	\label{sec:appendix_config}
	
	All experiments were conducted on a server with 4x NVIDIA A100 (80GB) GPUs. Total GPU hours: approximately 120 hours.
	
	\paragraph{Runtime Analysis.} Per-image perturbation generation takes $45.2 \pm 3.8$ seconds (mean$\pm$std) on a single A100 GPU, broken down as: forward pass through surrogates (18s), backward pass and gradient computation (22s), projection and clipping (5s).
	
	\paragraph{Victim Models:}
	\begin{itemize}
		\item \textit{Mobile-Agent-v2:} Uses Qwen-VL-Chat as the backbone via HuggingFace \texttt{transformers} (version 4.37.0).
		\item \textit{AppAgent:} Uses LLaVA-1.6-Vicuna-7B with the official codebase.
		\item \textit{Black-box APIs:} GPT-4o (gpt-4o-2024-05-13), Gemini-1.5-Pro (gemini-1.5-pro-001), Claude-3.5-Sonnet (claude-3-5-sonnet-20240620).
	\end{itemize}
	
	\paragraph{Attack Hyperparameters:}
	\begin{itemize}
		\item $\epsilon = 8/255$, $N = 50$ iterations, $\mu = 0.9$, $\alpha = 1/255$
		\item DI-FGSM Probability: $p=0.5$, resize range: 0.9-1.1
		\item Decoupling Weight: $\lambda = 1.5$, Repulsion Weight: $\beta = 0.5$
		\item Softmax Temperature: $\tau = 0.1$ for discrete grid outputs
	\end{itemize}
	
	\section{E-ShopBench Construction}
	\label{sec:appendix_dataset}
	
	E-ShopBench consists of XML-based UI layouts and corresponding screenshots.
	\begin{itemize}
		\item \textbf{Categories:} Electronics (50), Home Goods (50), Fashion (50), Accessories (50).
		\item \textbf{Price Distribution:} ``Cheap'' items: \$10--\$50. ``Expensive'' items: \$200--\$2000.
		\item \textbf{Layout Variation:} List Views (40\%), Grid Views (40\%), Detail Pages (20\%).
		\item \textbf{Platform Distribution:} Amazon (80), eBay (60), Taobao (60).
	\end{itemize}
	
	\section{Visual Anchor Bank Details}
	\label{sec:appendix_anchor}
	
	\subsection{Anchor Selection Criteria}
	We construct the anchor bank $\mathcal{A}_{cheap}$ by sampling 500 images of low-cost items chosen to reduce direct product overlap with E-ShopBench while emphasizing generic value-associated visual cues:
	\begin{enumerate}
		\item \textbf{Price Range:} Items priced below \$20
		\item \textbf{Visual Simplicity:} Products with minimal branding, simple packaging
		\item \textbf{Product Separation:} Avoid direct overlap in exact products or screenshots used in E-ShopBench
	\end{enumerate}
	
	\subsection{Category Distribution}
	\begin{itemize}
		\item Office supplies (pens, notebooks, folders): 150 images
		\item Basic kitchenware (plastic containers, utensils): 150 images
		\item Generic household items (cleaning supplies, storage boxes): 200 images
	\end{itemize}
	
	\subsection{Visual Characteristics}
	These items typically feature: simple backgrounds, generic branding, standard materials (plastic vs. brushed metal), utilitarian design, and basic product photography.
	
\subsection{Embedding Computation}
We pre-compute embeddings using \textbf{CLIP-ViT-L/14} (OpenAI's official checkpoint, 428M parameters). For ensemble attacks, we additionally compute an encoder-specific centroid $\bar{e}_{cheap}^{(m)}$ for each surrogate model's visual encoder from the same anchor bank.

	\section{Impact of Instruction Phrasing}
	\label{sec:appendix_instruction}
	
	As a supplementary descriptive stress test, rather than a replacement for the main clear-constraint benchmark, we probed PriceBlind under three prompt styles:
	\begin{itemize}
		\item \textbf{Explicit Constraint:} "Buy the item strictly under \$50." $\rightarrow$ ASR: about 68\%
		\item \textbf{Implicit Constraint:} "I need a budget-friendly option." $\rightarrow$ ASR: about 91\%
		\item \textbf{Urgent/Vague:} "Get me the best deal ASAP." $\rightarrow$ ASR: about 95\%
	\end{itemize}
	
	These supplementary checks suggest that ASR is highest for \textbf{Implicit Constraints} and \textbf{Vague Prompts}, where the model relies more heavily on its own semantic interpretation. Even with \textbf{Explicit Constraints}, the attack remains materially effective in this stress-test setting. These prompt-style results are intended to illustrate qualitative trend differences and should not be interpreted as directly comparable replacements for the main E-ShopBench benchmark figures.
	
\end{document}